\pdfoutput=1

\documentclass[11pt]{article}

\usepackage[]{ACL2023}

\usepackage{times}
\usepackage{latexsym}
\usepackage{graphicx}
\usepackage[T1]{fontenc}
\definecolor{deepgreen}{rgb}{0,0.5,0}

\usepackage[utf8]{inputenc}
\usepackage{enumitem}
\usepackage{caption}
\usepackage{subcaption}
\usepackage{microtype}
\usepackage{hyperref}
\usepackage{inconsolata}
\usepackage{amsmath}
\usepackage{multirow}
\usepackage{booktabs, multirow} 
\usepackage{soul}

\usepackage[capitalize,noabbrev]{cleveref}
\usepackage{xcolor}

\usepackage[]{todonotes}
\newcommand{\jj}[2][inline]{\space\todo[color=red!60,#1,size=\tiny]{{JOSEF: #2}}}

\usepackage[normalem]{ulem} 
\usepackage{listings}
\def\XXX#1{\textcolor{red}{XXX #1}}

\def\repl#1#2{\textcolor{red}{XXX \sout{#1}}\textcolor{blue}{\uline{#2}}}

\title{Breeding Machine Translations: Evolutionary approach to survive and thrive in the world of automated evaluation}

\author{Josef Jon \and Ondřej Bojar \\
 Charles University, Faculty of Mathematics and Physics \\
Institute of Formal and Applied Linguistics \\
  \texttt{\{jon,bojar\}@ufal.mff.cuni.cz}}

\begin{document}
\maketitle
\begin{abstract}

We propose a genetic algorithm (GA) based method for modifying $n$-best lists produced by a machine translation (MT) system. 
 Our method offers an innovative approach to improving MT quality and identifying weaknesses in evaluation metrics. 
Using common GA operations (mutation and crossover) on a list of hypotheses in combination with a fitness function (an arbitrary MT metric), we obtain novel and diverse outputs with high metric scores. 
With a combination of multiple MT metrics as the fitness function, the proposed method leads to an increase in translation quality as measured by other held-out automatic metrics.
With a single metric (including popular ones such as COMET) as the fitness function, we find blind spots and flaws in the metric. This allows for an automated search for adversarial examples in an arbitrary metric, without prior assumptions on the form of such example. As a demonstration of the method, we create datasets of adversarial examples and use them to show that reference-free COMET is substantially less robust than the reference-based version. 
\end{abstract}

\section{Introduction}
Attaining good translation quality in machine translation (MT) arguably relies on good automatic metrics of MT quality.
Recently, a new generation of  evaluation metrics was introduced. These metrics are based on embeddings computed by large pretrained language models and human annotation scores. 
 The improvements in metric quality resulted in renewed interest in metric-driven translation hypothesis selection methods, like Minimum Bayes Risk (MBR) decoding \cite{GOEL2000115,kumar-byrne-2004-minimum}.

Our method relies on MBR decoding and the genetic algorithm (GA;  \citealp{fraser1957simulation,bremermann1958evolution,Holland:1975}. Through combinations and mutations of translations produced by an MT model, we search for optimal translation under a selected metric. This is a novel approach to generating translation hypotheses in NMT.

 We find that by combining neural and surface form-based metrics in a GA's fitness function, it is possible to create better quality translations than by simple reranking of the initial hypotheses (as evaluated by held-out metrics). It also allows the combination of multiple sources for the translation, for example, MT, paraphrasing models and dictionaries.

Another use-case for our method is the identification of weak points in MT metrics. Flaws and biases of the novel neural metrics are being studied, for example, by \citet{hanna-bojar-2021-fine}, \citet{amrhein-sennrich-2022-identifying}, \citet{alves-EtAl:2022:WMT1} or \citet{kanojia-etal-2021-pushing}. In summary, these metrics have low sensitivity to errors in named entities and numbers. Also, they are not sufficiently sensitive to changes in meaning and critical errors, like negations.

These previous works on deficiencies of the metrics mostly focus on analyzing the outputs of MT systems and looking for certain types of mistakes. Another approach they use is changing the outputs to introduce specific types of mistakes. In contrast, our approach aims to find translations with high scores on certain metrics automatically, by optimizing the candidate translations for a selected metric. We believe that through this more explorative approach, it is possible to find unexpected types of defects. 

In summary,  the main contribution of our work is a novel method for producing translations, which can be used to improve translation quality and analyze automatic MT evaluation metrics.\footnote{Source code at \url{https://github.com/cepin19/ga_mt}}


\section{Related work}
\paragraph{Automated MT evaluation}
The traditional automatic MT metrics are based on comparing a translation produced by an MT system to a human reference based on a string similarity. Popular choices are ChrF~\cite{popovic-2015-chrf} and BLEU~\cite{papineni-etal-2002-bleu}. Multiple shortcomings of these metrics are well known \cite{callison-burch-etal-2006-evaluating,bojar-etal-2010-tackling,freitag-etal-2020-bleu,mathur-etal-2020-tangled,zhang-toral-2019-effect,graham-etal-2020-statistical}. 

\paragraph{Neural MT metrics}
Novel, neural-based MT metrics were introduced recently. They address some of the deficiencies of the string-based methods, but possibly introduce new types of errors or blind spots: BERTScore~\cite{bert-score}, BARTScore~\cite{bart-score}, PRISM~\cite{thompson-post-2020-automatic}, BLEURT~\cite{sellam-etal-2020-bleurt}, COMET~\cite{rei-etal-2020-comet,rei-etal-2021-references,rei-etal-2022-searching}, YiSi~\cite{lo-2019-yisi}, RoBLEURT~\cite{wan-etal-2021-robleurt} or 
UniTE~\cite{wan2022unite}. 

Using a shared embedding space, these metrics better compare source, translated, and reference sentences. Their evaluation in WMT Metrics tasks \cite{mathur-etal-2020-results,freitag-etal-2021-results,freitag-EtAl:2022:WMT} and other campaigns \cite{kocmi-etal-2021-ship} demonstrate stronger agreement with human judgment. 

While their system-level performance has been scrutinized, their segment-level performance remains less explored. \citet{extrinsic} indicates these metrics are unreliable for assessing translation usefulness at segment level. However, we still try to optimize individual sentences for improved scores.

\paragraph{Deficiencies in metrics}
The closest work to ours is \citet{amrhein-sennrich-2022-identifying}. Authors use MBR decoding to find examples of high-scoring, but flawed translations in sampled model outputs. The conclusion is that the studied metrics are not sensitive to errors in numbers and in named entities (NE).
\citet{alves-EtAl:2022:WMT1} automatically generate  texts with various kinds of errors to test for sensitivity of MT metrics to such perturbations. \citet{sun-etal-2020-estimating} claim that current MT quality estimation (QE) models do not address adequacy properly and \citet{kanojia-etal-2021-pushing} further show that meaning-changing errors are hard to detect for QE.

\paragraph{Genetic algorithm}
Variations of the genetic algorithm and evolutionary approaches in general for very diverse optimization problems are being studied extensively for more than half a century \cite{fraser1957simulation,bremermann1958evolution,sastry2005genetic}.  

Nevertheless, work on the utilization of the GA in machine translation is scarce. \citet{echizen-ya-etal-1996-machine} use GA for example-based MT. \citet{zogheib2011genetic} present multi-word translation algorithm based on the GA. \citet{ameur2016genetic} employ GA in phrase-based MT decoding. In the context of neural machine translation, GA was used to optimize architecture and hyperparameters of the neural network \cite{nmt_ga,nmt_ga2}.

\paragraph{Minimum Bayes risk decoding}
Our implementation of the fitness function depends on  Minimum Bayes Risk (MBR) decoding \cite{GOEL2000115,kumar-byrne-2004-minimum}. This selection method has regained popularity recently as new, neural-based MT metrics emerged  \cite{comet_mbr,bleurt_mbr,muller-sennrich-2021-understanding,jon-popel-bojar:2022:WMT}.
\section{Proposed solution}
Our approach depends on two methods: Minimum Bayes Risk decoding and genetic algorithm.

\subsection{Genetic algorithm}
We propose the use of a GA to find new translation hypotheses. GA is a heuristic search algorithm defined by a \textit{fitness function}, operators for combination (\textit{crossover}) and modification (\textit{mutation}) of the candidate solutions, and a~\textit{selection} \textit{method}.

Before running the GA algorithm, an initial \textit{population} of a chosen number of candidate solutions is created. A single solution is called an \textit{individual}, and it is encoded in a discrete way (often as a list) by its forming units, \textit{genes}. The resulting representation of an individual is called a \textit{chromosome}. 
All chromosomes have the same length to simplify the corssover operation, but we add placeholders for empty tokens to account for additions, as discussed later. 

The algorithm itself consists of evaluating each solution in the population using the fitness function and stochastically choosing parent solutions for the new generation by the selection algorithm. Crossover is used on the chromosomes of the parents to create their offspring (\textit{children}). The mutation is used on the children and they form a new generation of the same size. This is repeated for a given number of iterations (\textit{generations}).

In our proposed method, the candidate solutions are translation hypotheses produced by an MT model. Genes are tokens and the mutation operation replaces, deletes, or adds a token in a chromosome. The eligible new tokens are chosen from a set of valid tokens. We discuss methods of construction of this set in \cref{sec:mut}.

To allow for variable lengths of the solutions and the add or delete operations, we add genes representing an empty string after each token gene, and all the candidates are also right-padded with the empty string genes. The final length of all the candidates is equal to the length of the longest candidate multiplied by a constant  $k$. The empty string genes can be mutated to a non-empty gene, which is equivalent to inserting a new token into the candidate. Inversely, a non-empty string  gene can be mutated to an empty string gene, which is equivalent to removing a token. Empty genes have no influence on the fitness score. Below we show the encoding of two translation hypotheses
for $k=1.1$:

\scriptsize
\begin{lstlisting}[language=Python,stringstyle=\color{deepgreen}]
sent1=['Genetic','','algorithm','','can','','be','','used','',
'to','' ,'produce','','novel','','solutions','','.','','','']}

sent2=['Genetic','','algorithm','','creates','','new','',
'solutions','','.','','','','','','','','','']}
\end{lstlisting}
\normalsize

\begin{figure*}[h]
\centering
\includegraphics[width=0.9\linewidth]{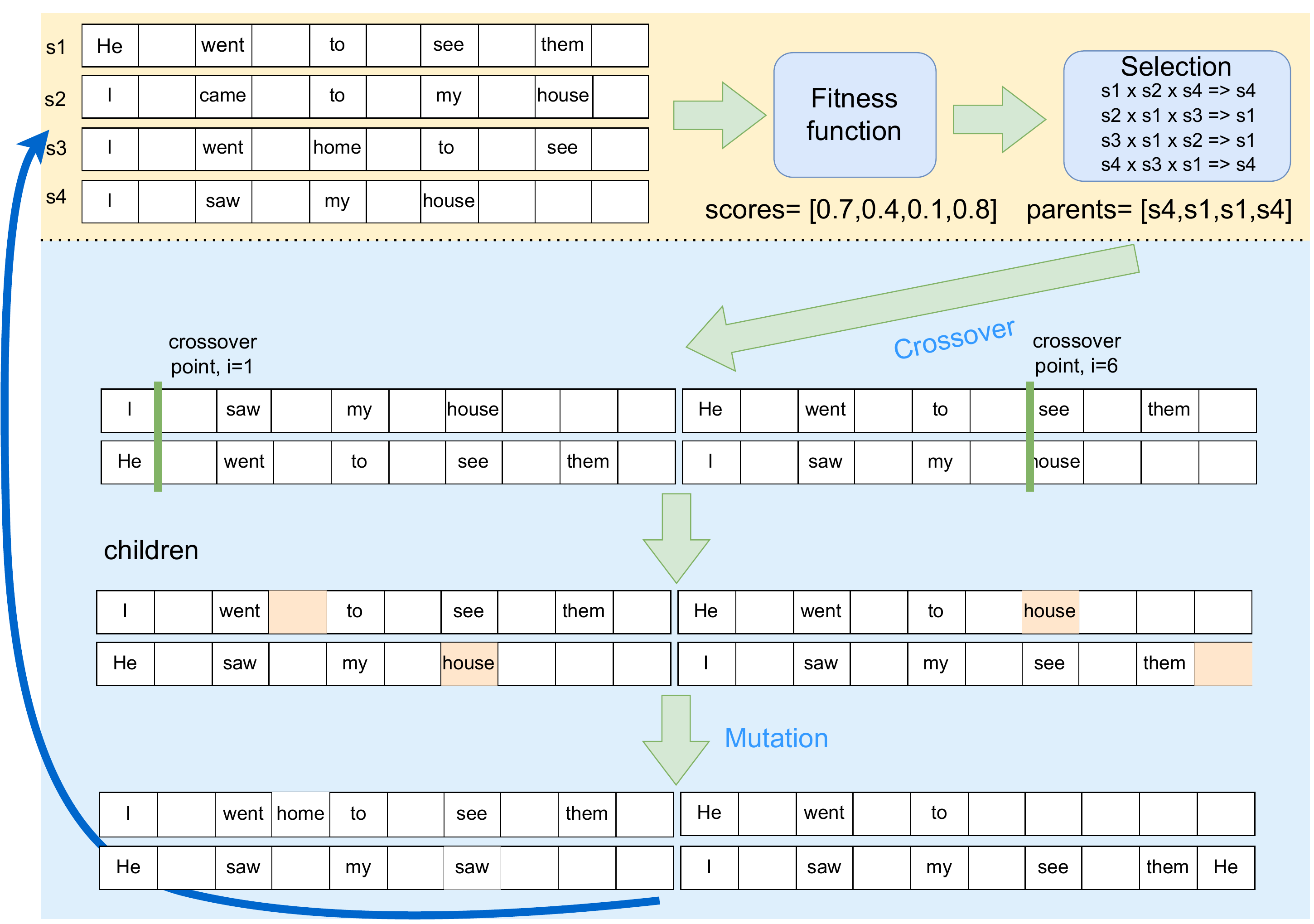}
\caption{One iteration of the GA algorithm for a population of 4 individuals. The steps with a yellow background are equivalent to simple reranking, the steps with blue background introduce the operations of the genetic algorithm.}
\label{fig:ga}
\end{figure*}

\paragraph{Fitness function}
Fitness functions are MT evaluation metrics, see \Cref{sec:exp}. For some of the experiments, the fitness function is composed of multiple metrics. In that case, the scores are simply summed -- we did not explore scaling them or using multi-objective GA \cite{murata1995moga,surry1997comoga,gao2000study,deb2002fast}.
\paragraph{Selection}
To select parents for the new generation, we use tournament selection with $n=3$. 
For each individual in the population, two other individuals are randomly chosen and the one with the best value of the fitness function out of the three is selected as one of the parents for a new generation. \cref{fig:ga} illustrates this, including the fact that many individuals can be selected repeatedly through this process.
\paragraph{Crossover operation}
We iterate through the parents by pairs, each pair is crossed-over with probability $c$. A random index $i$ in a chromosome is selected and two children are created, the first one inherits the part of chromosome up to $i$ from the first parent and the part from $i$ from the second parent and vice-versa for the second offspring. For parents \texttt{p1} and \texttt{p2} and children \texttt{c1} and \texttt{c2}: 

\begin{lstlisting}[language=Python,stringstyle=\color{deepgreen}]
c1=p1[:i]+p2[i:]; c2=p2[:i]+p1[i:]
\end{lstlisting}

\paragraph{Mutation operation}
The children produced by the cross-over operation are mutated. Each gene (token) is mutated with a probability $m$. Mutation replaces the token (or empty string placeholder) with a randomly selected one from the set of all possible tokens. This set also includes empty string placeholder, which is equivalent to token deletion. The approaches to the construction of this set are described in \cref{sec:mut}. After the mutation phase, the new generation is ready and the next iteration of GA is performed. One iteration of the whole GA process is illustrated in \cref{fig:ga}.

\paragraph{MT Metrics and Fitness vs. Evaluation }

Optimizing the word composition of a translation towards an arbitrary metric is subject to Goodhart's law -- once a metric is used as a goal to optimize towards, it ceases to be a good measure of final quality \cite{Strathern1997}. Thus, we cross-evaluate with held-out metrics not used for optimization (even though these metrics might still be linked with the optimization metrics by spurious correlations caused by similar metric design, model architecture, or training data).  We search for adversarial examples for the specific metrics, i.e. translations scoring high in the objective metric, but low in held-out metrics. This can be used to create training sets of negative examples. We use ChrF, BLEU, wmt20-comet-da~\cite{rei-etal-2020-comet}, wmt20-comet-qe-da-v2 as the objective metrics and wmt21-comet-mqm, eamt22-cometinho-da, BLEURT~\cite{sellam-etal-2020-bleurt} and UniTE~\cite{wan-etal-2022-unite} as the held-out metrics.
\addtolength{\belowcaptionskip}{-2.8mm}

\subsection{MBR decoding}

NMT models predict a probability distribution over translations for a given source sentence. A common method for selecting a final translation given this distribution is known as ``maximum-a-posteriori" (MAP) decoding. Because of the computational complexity of exact MAP decoding, approximations such as beam search \cite{koehn-etal-2003-statistical} are used. Many limitations of MAP were described recently \cite{stahlberg-byrne-2019-nmt,meister-etal-2020-beam} and other approaches were proposed.

One of the alternatives is MBR decoding. It is a decision rule that selects the translation based on a value of a utility function (and thus minimizes expected loss, or \textit{risk}) rather than model probability. MT metrics are often used as utility functions. In an ideal case, we have a distribution $p(y|x)$ over all possible correct translations $y$ of source sentence $x$ available, which is not the case in real-world scenarios.  Given the space of all possible target language sentences $\mathcal{H}{(x)}$ and utility function $\mathcal{U}$, we search for the optimal translation $h^*$:
\addtolength{\abovedisplayskip}{-1.8mm}
\addtolength{\belowdisplayskip}{-1.3mm}
$$h^* = argmax_{h \in  \mathcal{H}(x)} E_{p(y|x)}[\mathcal{U}(y, h)]$$
A fixed number of translation hypotheses produced by the MT model can be used as an approximation of the reference translations distribution $p(y|x)$ in practice. Still, the number of possible hypotheses $\mathcal{H}{(x)}$ is infinite -- it consists of all conceivable sentences in the target language. For this reason, the same set of translations as for references is also used as candidate hypotheses. This leads to an implementation where MBR decoding can be seen as consensus decoding -- a translation 
that is the most similar to all the other translations in the set is selected. Some of the recent embedding-based metrics also take the source sentence into account. In that case, utility is defined as $\mathcal{U}(x, y, h)$. In such cases, the process is no longer equivalent to consensus decoding due to the influence of the source.

\section{Experiments}
This section describes our experimental setup and results. We compare reranking of $n$-best lists to the application of the GA on them.
\label{sec:exp}

\subsection{Data}
We trained Czech-English MT model on CzEng 2.0~\citep{kocmi2020announcing}, a mix of parallel data (61M) and Czech monolingual data back-translated into English (51M). For experiments with dictionaries, we use a commercial Czech-English dictionary. We use \texttt{newstest-19}~\cite{barrault-etal-2019-findings} as the dev set and \texttt{newstest-18}~\cite{bojar-etal-2018-findings} as the test set. Due to the high computational requirements of our approach, we only evaluate the first 150 sentences from the test set in all the experiments. We call this test set \texttt{newstest-18-head150}. We used a commercial lemmatizer.\footnote{\url{http://www.lingea.com}} for lemmatization and word form expansion performed in some of the experiments, 
We tokenize the data into subwords with SentencePiece~\cite{kudo-richardson-2018-sentencepiece} and FactoredSegmenter.\footnote{\url{https://github.com/microsoft/factored-segmenter}}

\subsection{Model}
We train \texttt{transformer-big} using MarianNMT~\cite{mariannmt} with default hyperparameters.
\subsection{Hardware}
We ran all the experiments on a grid server with heterogeneous nodes, with Quadro RTX 5000, GeForce GTX 1080 Ti, RTX A4000, or GeForce RTX 3090 GPUs. The running time depends on population size, number of generations, and fitness function. We leave the first two fixed, so the computational requirements are most influenced by the fitness function. For the most computationally intensive fitness (combination of \texttt{wmt20-comet-da} and \texttt{wmt20-comet-qe-da-v2}), optimizing 150 examples on RTX A4000 takes 5 days. We discuss the computational requirements in \cref{sec:limitations}.

\subsection{Metrics}
We abbreviate some of the longer metrics' names further in the text in order to save space.\footnote{CMT20 (wmt20-comet-da), CMT21 (wmt21-comet-mqm), CMTH22 (eamt22-cometinho-da), QE (wmt20-comet-qe-da-v2), BLEURT (BLEURT-20), UniTE (UniTE-MUP)}

 For BLEU and ChrF we use SacreBLEU~\cite{post-2018-call}. We use $\beta=2$ for ChrF in all the experiments (i.e. ChrF2). For COMET\footnote{\url{https://github.com/Unbabel/COMET}}, BLEURT\footnote{\url{https://github.com/google-research/bleurt}} and UniTE\footnote{\url{https://github.com/NLP2CT/UniTE}} scores we use the original implementations. We use paired bootstrap resampling \cite{koehn-2004-statistical} for significance testing.

\subsection{GA parameters}
We did not search for optimal values of GA parameters due to high computational costs.
The initial population is formed by 
20-best hypotheses obtained by beam search and 20 sampled ones, copied 50 times over to obtain a population size of 2000.
We select parents for the new generation with tournament selection ($n=3$) and then we combine them using a crossover rate $c=0.1$. The mutation rate for the mutation of non-empty genes to different non-empty genes $m$ is $1/l$, where $l$ is the chromosome length. For mutating empty to non-empty gene (word addition) or vice-versa (deletion), the rate is $m/10$. We run 300 generations of the GA.  
\subsection{Possible mutation sources}
\label{sec:mut}
We consider three possible sources for the mutation tokens set, i.e. the set of tokens that can replace another token in the chromosome: 
\begin{enumerate}[label=\arabic*)]
\item \textit{init} -- set of all the tokens from the initial population (only tokens that are present in initial hypotheses can be used for the optimization).
\item \textit{dict} -- we performed word-by-word dictionary translation of each the source sentence, resulting in a set of English tokens. The source sides of the dictionary and the source sentence are lemmatized for the search, and target token forms are expanded to cover all surface forms.
\item \textit{wordlist} -- all words from an English wordlist.\footnote{\url{https://github.com/dwyl/english-words}}
\end{enumerate}

\subsection{Results}
\begin{table*}[!htb]\centering
\scriptsize
\begin{tabular}{lclrrrrrrrrr}\toprule
\textbf{Source} &\textbf{Rerank} &\textbf{Metric} &\textbf{ChrF} &\textbf{BLEU} &\textbf{CMT20} &\textbf{CMT21} &\textbf{CMTH22} &\textbf{QE} &\textbf{BLEURT} &\textbf{UniTE} \\\midrule

beam 5 &- &log-prob &56.4 &28.9 &0.4995 &0.0399 &0.5025 &0.2472 &0.7066 &0.3004 \\
\midrule
\multirow{7}{*}{beam 20} &- &log-prob &56.7 &30.1 &0.5007 &0.0399 &0.5017 &0.2477 &0.7078 &0.3018 \\
\cmidrule{2-11}

&\multirow{3}{*}{Oracle}  &ChrF &\sout{64.1} &40.3 &0.6046 &0.0423 &0.6552 &0.2592 &0.7449 &0.3953 \\
& &BLEU &63.0 &\sout{41.1} &0.5897 &0.0419 &0.6434 &0.2573 &0.7390 &0.368 \\
& &CMT20 &62.0 &37.7 &\sout{0.6903} &0.0431 &0.6875 &0.2949 &0.7551 &0.4641 \\
\cmidrule{2-11}
&\multirow{3}{*}{MBR} & ChrF &\sout{57.1} &30.4 &0.5162 &0.0399 &0.5105 &0.2514 &0.7075 &0.3056 \\
 & &BLEU &56.3 &\sout{29.6} &0.5102 &0.0399 &0.5104 &0.2357 &0.7079 &0.2958 \\
& &CMT20 &56.8 &30.6 &\sout{0.5686} &0.0404 &0.5281 &0.2818 &0.7160 &0.3313 \\
\midrule
\multirow{7}{*}{sampled 20} &- &log-prob & 53.0 &25.5 &0.3557 &0.0371 &0.3878 &0.1350 &0.6661 &0.1277 \\
\cmidrule{2-11}

&\multirow{3}{*}{Oracle}& ChrF &\sout{62.5} &37.1 &0.4848 &0.0392 &0.5346 &0.1471 &0.7007 &0.2211 \\
 & &BLEU &60.5 &\sout{39.6} &0.4143 &0.0382 &0.4806 &0.1133 &0.6872 &0.1609 \\
& &CMT20 &58.0 &31.7 &\sout{0.6630} &0.0419 &0.6313 &0.2526 &0.7336 &0.4061 \\
\cmidrule{2-11}

&\multirow{3}{*}{MBR}&  ChrF & \sout{55.4} &28.2 &0.4376 &0.0386 &0.4621 &0.2017 &0.6926 &0.2274 \\
 & &BLEU & 54.3 &\sout{28.2} &0.3998 &0.0381 &0.4493 &0.1713 &0.6855 &0.1892 \\
& &CMT20 &54.4 &28.0 &\sout{0.5515} &0.0403 &0.5194 &0.2617 &0.7062 &0.2931 \\
\midrule
\multirow{9}{*}{\shortstack{beam 20 \\ + \\ sampled 20}} & -&  log-prob&56.6 & 30.1&  0.5002	 & 0.0399	& 0.5044	 & 0.2436 & 0.7067 & 0.3001	 \\
\cmidrule{2-11}

&\multirow{3}{*}{Oracle} &ChrF &\sout{\textit{65.4}} &41.9 &0.5973 &0.0417 &0.6448 &0.2330 &0.7395 &0.3818 \\
 & &BLEU &63.7 &\sout{\textit{43.2}} &0.5507 &0.0410 &0.6100 &0.2205 &0.7286 &0.3236 \\
& &CMT20 &61.9 &37.6 &\sout{\textit{0.7154}} &\textit{0.0433} &\textit{0.7017} &\textit{0.2872} &\textit{0.7561} &\textit{0.477} \\
\cmidrule{2-11}

&\multirow{5}{*}{MBR}& ChrF &\sout{56.9} &30.3 &0.5192 &0.0399 &0.5112 &0.2517 &0.7092 &0.3059 \\
 & &BLEU &56.4 &\sout{30.0} &0.5047 &0.0398 &0.5100 &0.2403 &0.7069 &0.2958 \\
& &CMT20 &57.4 &\textbf{31.2} &\sout{0.5853} &0.0409 &0.5390 &0.2930 &0.7193 &0.3413 \\
& & QE & 55.7 & 29.5 & 0.539 & 0.0412 & 0.4976 & \sout{\textbf{0.3841}}  &0.7140 & 0.3274 \\
& &CMT20+QE+BLEU &\textbf{57.5} &\sout{\textbf{31.2}} &\sout{\textbf{0.5983 }}&\textbf{0.0417 }&\textbf{0.5596 }&\sout{0.3620} &\textbf{0.7255} &\textbf{0.3686}\\
\bottomrule
\end{tabular}
\caption{Results of baseline translations and their reranking by multiple metrics on \texttt{newstest-18-head150}.  Higher is better for all the metrics. The best scores for MBR-based reranking are shown in bold, the best scores for reference-based reranking are written in italics.
 We strike out the values where the same metric was used to rerank and also evaluate the outputs.}
\label{tab:base}
\end{table*}

\paragraph{Reranking}
We translated \texttt{newstest-18} by the baseline model using beam search with beam size 20. We also sampled another 20 translation hypotheses for each source sentence from the model. We rerank these lists by BLEU, ChrF and CMT20 metrics in two manners: either with knowledge of the true manual reference (i.e. oracle) or using MBR decoding. GA is not used in these experiments. There are two ways of using multiple references with BLEU: either compute single-reference scores for all the references separately and average them or use the multi-reference formula. We use the former.

The results are presented in  \cref{tab:base}. The confidence ranges are shown in \cref{app:sig}, \cref{tab:base_sig}. The 1st column shows the origin of the hypotheses.\footnote{The outputs produced with beam size 5 are not used in further experiments, they are shown for comparison to account for the beam search curse (larger beam sizes sometimes result in worse translation outputs, \citealp{koehn-knowles-2017-six}).} The 2nd column shows if the reference was used for reranking (\textit{Oracle}), or the other hypotheses and MBR decoding were used instead (\textit{MBR}). No reranking \hbox{(-)} means that the candidate with the highest model's length-normalized log-prob is evaluated. The 3rd column indicates which metric was used for the reranking (the objective function). The remaining columns are the values of the evaluation metrics (computed with respect to the reference).

For most of the metrics, MBR-reranked hypotheses outperform the log-prob baseline, even though by a smaller margin than the reference-reranked (oracle) ones. In some cases, optimizing with MBR towards one metric leads to a deterioration of scores in other metrics.
The metrics most prone to this problem are QE, ChrF and BLEU. MBR rescoring with QE results in worse ChrF, BLEU and CMTH22 scores than the baseline, suggesting this metric is unsuitable for such application. CMT20 and especially the combination of CMT20+QE+BLEU are more robust, with the latter improving in all the metrics over the baseline. As shown further, both the negative and positive effects are more pronounced with GA. 
Reranking with knowledge of the reference is unsurprisingly performing better than MBR reranking. Here, we use it to show the upper bound of improvements attainable by reranking. In further experiments, reference-based GA is also used to analyze the objective metrics. 

We also notice that while reranking beam search results leads to better final outcomes than reranking sampling results, a combination of both provides the best scores. All further experiments start with a population consisting of this combination of both.

\begin{table*}[!htp]\centering
\scriptsize
\begin{tabular}{cllrrrrrrrrrr}\toprule
\textbf{Fitness} &\textbf{Mut} &\textbf{\#runs} &\textbf{ChrF} &\textbf{BLEU} &\textbf{CMT20} &\textbf{CMT21} &\textbf{CMTH22} &\textbf{QE} &\textbf{BLEURT} &\textbf{UniTE} &\textbf{new} \\\midrule
\multirow{4}{*}{ChrF} &- &9 &\sout{71.4} &48.3 &0.4144 &0.0369 &0.5493 &0.0104 &0.6853 &0.2018 &0.79 \\
&init &9 &\sout{84.9} &60.0 &0.0994 &0.0308 &0.3300 &-0.2777 &0.6266 &-0.0617 &0.92 \\
&init+dict &9 &\sout{87.1} &58.0 &0.0813 &0.0304 &0.3171 &-0.3004 &0.6360 &-0.0784 &0.93 \\
&wordlist & 1 &\sout{83.2}&48.5 &-0.3729 &0.0214 &-0.2245 &-0.4932 &0.5525 &-0.5097 &0.93\\
\midrule

\multirow{4}{*}{BLEU} &- &9 &68.0 &\sout{50.8}&0.4016 &0.0374 &0.5182 &0.0299 &0.6779 &0.1698 &0.76 \\
&init &9 &77.6 &\sout{68.9} &0.2693 &0.0353 &0.4747 &-0.1663 &0.6605 &0.0636 &0.92 \\
&init+dict &9 &79.6 &\sout{69.5} &0.2691 &0.0350 &0.4865 &-0.1866 &0.6631 &0.0627 &0.93 \\
&wordlist  &1 &68.3 &\sout{54} &-0.0306 &0.0292 &0.1243 &-0.3014 &0.5727 &-0.2492 &0.91\\
\midrule

\multirow{4}{*}{CMT20} & - &1 &64.6 &40.4 &\sout{0.7724} &0.0441 &0.7593 &0.2981 &0.7619 &0.5141 &0.67 \\

&init & 1 &70.1 &49.2 &\sout{0.8874}&0.0462 &0.868 &0.2476 &0.7763 &0.5824 &0.91 \\
&init+dict &6 &69.2 &46.3 &\sout{0.8974} &0.0467 &0.8897 &0.2598 &0.7790 &0.5876 &0.92 \\
&wordlist &1 &64.5 &41.1 &\sout{0.8371} &0.0446 &0.736 &0.2656 &0.7453 &0.4743 &0.87 \\
\bottomrule
\end{tabular}
\caption{Scores of translations on \texttt{newstest-18-head150} created by GA with the knowledge of the reference for the fitness function.  Higher is better for all the metrics.  Striked-out scores indicate results where fitness and evaluation metric coincide. 
}\label{tab:ga_ref}
\end{table*}

\paragraph{Genetic algorithm}
We use the same metrics for GA fitness function as for reranking. Experiments were again conducted with either the knowledge of the reference or with MBR decoding. The results for GA with reference are presented in \cref{tab:ga_ref} (confidence ranges in \cref{app:sig},S \cref{tab:ga_ref_sig}).  The first two columns indicate the metric used as the fitness function and the source of the possible tokens for the mutation. The third column shows how many runs were averaged to obtain the mean scores shown in the remaining columns. The last column shows the ratio of the final selected hypotheses that were not in the initial pool produced by the MT model, but were created by GA operations.

We see that the GA can optimize towards an arbitrary metric better than simple MBR reranking. For example, the best ChrF score for GA is 87.1 compared to 65.4 for reranking. 
The results also suggest that the string-based metrics (ChrF and BLEU) are  prone to overfitting -- translations optimized for these metrics score poorly in  other  metrics. CMT20 is more robust -- we see improvements over the baseline in all the metrics after optimization for CMT20. 


\Cref{tab:ga_mbr} presents the results of the experiments aimed to improve the translation quality (confidence ranges for the scores are in \cref{app:sig}, \cref{tab:ga_mbr_sig}). The reference is not provided  and MBR decoding (always computed with regard to the initial population) is used instead. This way, it is feasible to use the approach to improve translations in a real-world scenario with no reference. 
We measure the improvement by held-out metrics.\footnote{CMT21, CMTH22, BLEURT and UniTE} We consider UniTE to be the most trustworthy. It was created most recently and some of the flaws of the other metrics were already known and mitigated. It also correlates well with human evaluation~\cite{freitag-EtAl:2022:WMT} and it is developed by a different team than the COMET metrics, which slightly decreases the chances for spurious correlations of the scores not based on translation quality.

The metrics that only compare the translation with a reference (BLEU, ChrF) without access to the source sentence do not perform well as a fitness function. Since MBR decoding in such cases works as a consensus decoding, i.e. the most similar candidate to all the others has the best fitness, there is no evolutionary pressure to modify the individuals. 

Optimizing for QE or ChrF results in a large decline in scores for other metrics. These metrics are prone to scoring malformed, nonsensical or unrelated sentences well. This is  analyzed in \Cref{sec:analysis}.
The sum of QE, CMT20 and BLEU as the fitness function reaches the best score in UniTE and does not show significant degradation in other metrics.

The ratio of examples where held-out scores improve, decrease or do not change after GA is shown in  \Cref{tab:ratio}. We compare the scores both to log-prob selected hypotheses and MBR reranked ones.
We again see that the combination of CMT20+QE+BLEU performs best. GA with the individual metrics as the fitness function leads more often to a decrease than an increase of held-out metrics compared to reranking. This suggests the effect of GA on the translation quality is negative if the fitness function is not chosen well.

\begin{table}[!h]\centering

\scriptsize
\begin{tabular}{lccc}\toprule
Fitness  & +   & - & = \\\midrule
BLEU  & 22\%/1\%&29\%/7\%&49\%/92\% \\
CHRF   & 13\%/1\%&69\%/65\%&18\%/33\%  \\
CMT20 & \textbf{54\%}/23\% &\textbf{ 39\%}/32\% &\textbf{7\%}/45\% \\
CMT20+QE+BLEU & \textbf{62\%}/\textbf{43\%} & \textbf{35\%}/\textbf{35\%} & \textbf{3\%}/\textbf{23\% }\\
\bottomrule
\end{tabular}
\caption{Percentage of examples from \texttt{newstest-18-head150} where the held-out score (UniTE) improves (2nd column), degrades (3rd column), or doesn't change (4th column) for GA compared to log-prob selection/MBR reranking. The first column shows which metric was used as the fitness function. Bold results are the ones where held-out scores improve for more examples rather than where they deteriorate.} \label{tab:ratio}
\end{table}

\begin{table*}[!htp]\centering
\scriptsize
\begin{tabular}{lllrrrrrrrrrr}\toprule
\textbf{Fitness} &\textbf{Mut} &\textbf{\#runs} &\textbf{ChrF} &\textbf{BLEU} &\textbf{CMT20} &\textbf{CMT21} &\textbf{CMTH22} &\textbf{QE} &\textbf{BLEURT} &\textbf{UniTE} &\textbf{new} \\ \midrule
baseline & - & -&56.6 & 30.1&  0.5002	 & 0.0399	& 0.5044	 & 0.2436 & 0.7067 & 0.3001 &0.00 \\ 
best rerank & - & - & 57.5 & 31.2 & 0.5983 &0.0417 & \textbf{0.5596} &0.3620 & \textbf{0.7255 }& 0.3686 & 0.00 \\
\midrule
\multirow{4}{*}{ChrF} &- &7 &\sout{57.2} &30.0 &0.4769 &0.0387 &0.4877 &0.2140 &0.6963 &0.2549 &0.26 \\
&init &5 &\sout{\textbf{57.9}}&27.1 &0.2197 &0.0336 &0.2717 &0.0047 &0.5979 &0.0211 &0.73 \\
&init+dict &5 &\sout{\textbf{57.9}} &27.8 &0.2529 &0.0342 &0.2952 &0.0198 &0.6095 &0.0439 &0.68 \\
&wordlist & 1 &\sout{57.5} &29.4 &0.3614 &0.0365 &0.3949 &0.1343 &0.6558 &0.1214 &0.45 \\
\midrule

\multirow{4}{*}{BLEU} &- &9 &56.4 &\sout{30.0} &0.4997 &0.0397 &0.5066 &0.2366 &0.7059 &0.2901 &0.04 \\
&init &7 &56.4 &\sout{29.9} &0.5004 &0.0396 &0.5071 &0.2322 &0.7039 &0.2850 &0.09 \\
&init+dict &6 &56.3 &\sout{29.8} &0.5001 &0.0396 &0.5068 &0.2320 &0.7039 &0.2847 &0.08 \\
&wordlist&1& 56.3 &\sout{29.8} &0.4986 &0.0396 &0.5052 &0.2332 &0.7042 &0.2853 &0.07 \\
 \midrule

\multirow{4}{*}{CMT20} &- & 1 &57.6 &\textbf{31.7} &\sout{0.5988} &0.0410 &0.5385 &0.2939 &0.7192 &0.3446 &0.24 \\
&init & 1 &56.2 &28.4 &\sout{0.6247} &0.0410 &0.5382 &0.2893 &0.7177 &0.3366 &0.52  \\
&init+dict &5 &56.7 &29.4 &\sout{0.6188} &0.0411 &0.5412 &0.2880 &0.7124 &0.3362 &0.49 \\
&wordlist & 1 &57.3 &31.1 &\sout{0.6012} &0.041 &0.5288 &0.2907 &0.7162 &0.3385 &0.28  \\
\midrule

\multirow{2}{*}{QE}
&init+dict &1 &45.5 &13.2 &0.3353 &0.0398 &0.1836 &\sout{\textbf{0.5554}}&0.6018 &0.0324 &0.99\\
&wordlist &1 &46.0 &16.7 &0.1207 &0.0368 &-0.0643 &\sout{0.5514} &0.5349 &-0.3264 &0.99  \\
\midrule

\multirow{2}{*}{QE+CMT20}
&init &4 &55.0 &24.3 &\sout{\textbf{0.6387}} &\textbf{0.0431} &0.5066 &\sout{0.4778} &0.6963 &0.3444 &0.86 \\
&init+dict &5 &54.5 &24.4 &\sout{0.6321} &\textbf{0.0430} &0.5038 &\sout{0.4797} &0.6973 &0.3477 &0.85 \\
\midrule

\\ 

\multirow{2}{*}{QE+CMT20+BLEU} 
&init &1 &57.5 &\sout{29.5} &\sout{0.6266} &\textbf{0.0429} &0.5403 &\sout{0.4198} &0.7174 &\textbf{0.3946} &0.70 \\
&init+dict &3 & 57.4 &\sout{29.9} &\sout{0.6254} &\textbf{0.0429} &0.5403 &\sout{0.4180} &0.7169 &0.3916 &0.65 \\
\bottomrule
\end{tabular}
\caption{ Scores of translations on \texttt{newstest-18-head150} created by GA \textbf{without} knowledge of the
reference in the fitness function, using other hypotheses and MBR decoding instead. For better comparison we reiterate the baseline and best MBR reranking results (equivalent to GA with a single generation) in the first two rows. Higher is better for all the metrics. The best scores for MBR-based GA are shown in bold, for reference-based reranking in italics. Results where fitness and evaluation metrics coincide are striked out.}\label{tab:ga_mbr}
\end{table*}

\section{Analysis}
\label{sec:analysis}
In this section, we analyze the GA procedure and the behavior of evaluation metrics. 
\subsection{GA process}
\begin{figure}[h]
\addtolength{\belowcaptionskip}{3.1mm}
\centering
\begin{subfigure}[b]{1.00\linewidth}
\caption{}
\includegraphics[width=1.00\linewidth]{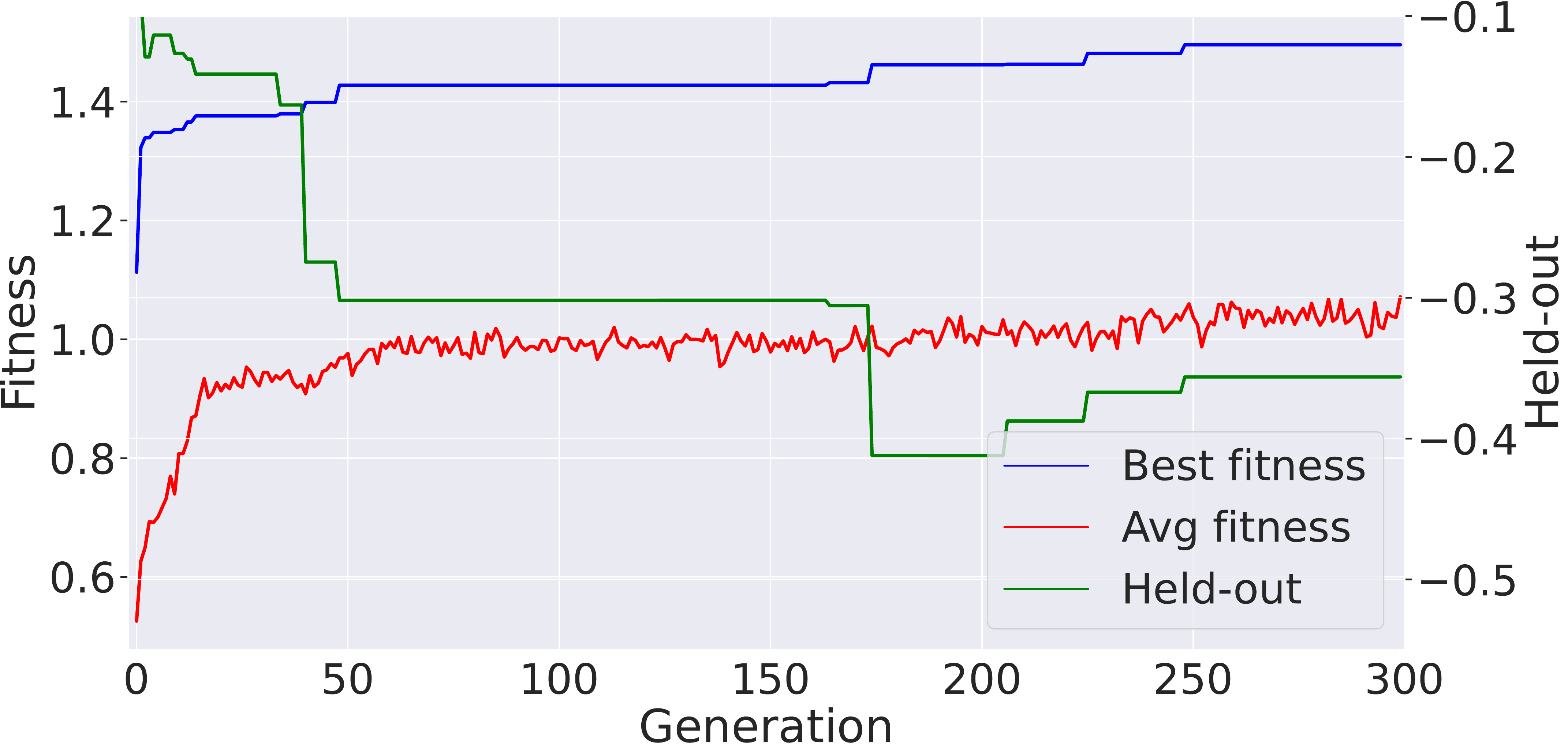}
\end{subfigure}

\begin{subfigure}[b]{1.00\linewidth}
\caption{}
\includegraphics[width=1.00\linewidth]{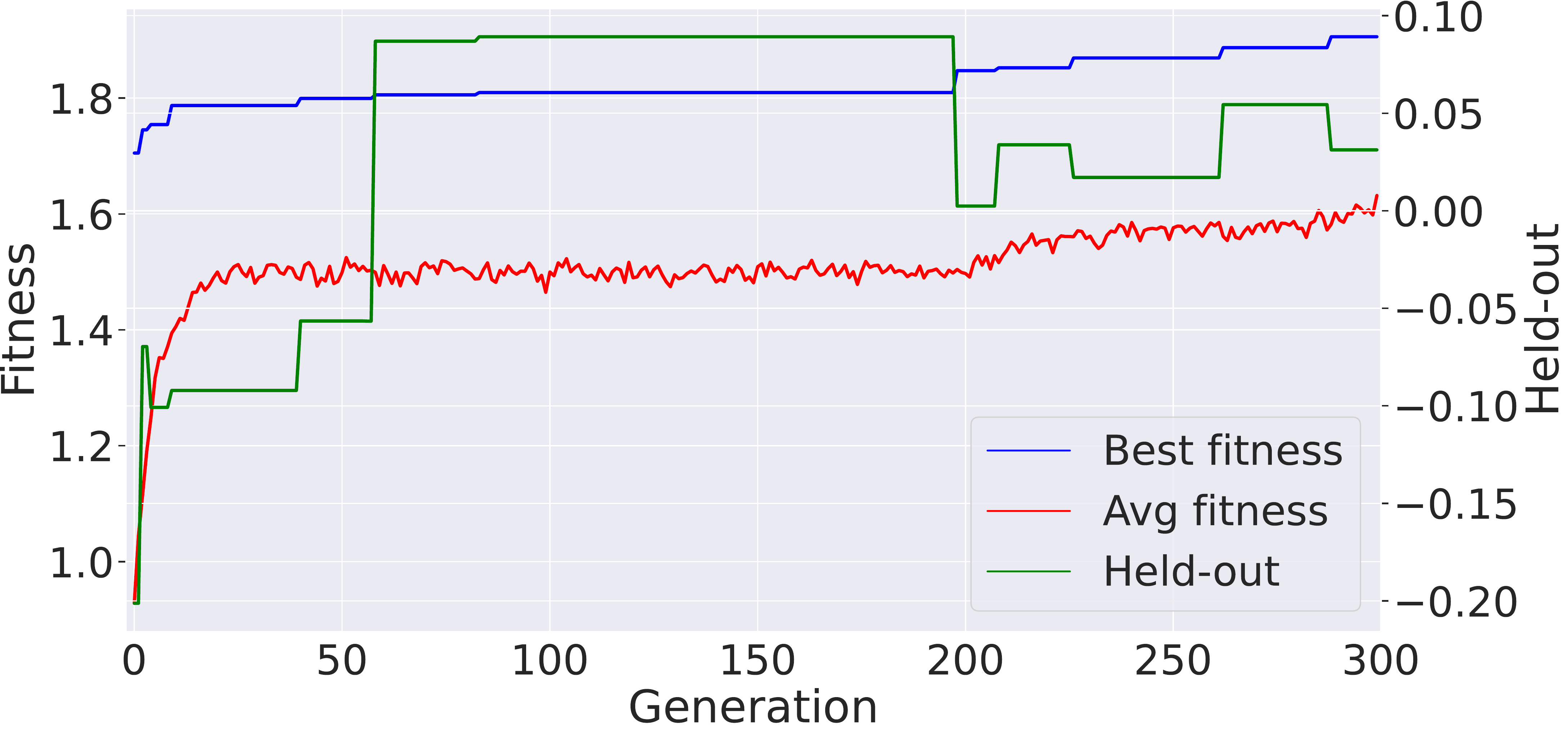}
\end{subfigure}
\addtolength{\belowcaptionskip}{-3.1mm}
\caption{Behavior of best and population average fitness, compared to held-out metric score $H$ of the best solution during a run of GA for two selected examples. $H$ (UniTE) does not correlate well with the fitness metric (CMT20+QE+BLEU) and the GA is detrimental from the point of view of the $H$ in Example a). In Example b), $H$ behaves similarly to the fitness function.}
\label{fig:behavior}
\end{figure}


\paragraph{Fitness vs. held-out metric} We analyzed the behavior of the average fitness function over the whole population, best solution fitness, and held-out metric score during the  GA process using CMT20+QE+BLEU as the fitness and UniTE as the held-out metric (\Cref{fig:behavior}). Results show GA consistently improved fitness values from initial solutions and increased average fitness. However, the correlation between fitness and held-out metrics varied: Example \textit{a)} shows a decrease in final held-out score despite improved fitness, while Example \textit{b)} shows aligned increases in both scores. \Cref{tab:ratio} suggests case \textit{b)} is more typical in our test set.

\subsection{Search for adversarial examples}

\begin{table}[!htp]\centering
\scriptsize
\begin{tabular}{lcc}\toprule
$O$ &$O_{init}+m_o<O_{ga}$ & ... $ \land H_{init}>H_{ga}+m_h$ \\\midrule
CMT20 &128 (85\%) &57 (38\%) \\
QE & 148 (99\%) & 142 (95\%) \\
BLEU &150 (100\%) & 113 (75\%) \\
\bottomrule
\end{tabular}
\caption{Number of examples from \texttt{newstest-18-head150} which improved in optimization metric after GA (2nd column) and at the same time deteriorated in held-out metric (3rd column)} \label{tab:adv_counts}
\end{table}

As a radically different goal, we use  GA  to search for examples that score high in the fitness function but are evaluated poorly by held-out metrics. This allows us to find blind spots in specific metrics without previous assumptions about the type of errors that could be ignored by the given metric. Such adversarial examples are defined as follows: for each test set example \textit{e}, we compute the scores of the hypotheses produced by the MT model using both the optimization metric $O$ and the held-out metric $H$.  We rank the hypotheses by $O$. The scores of the best hypothesis are referred to as $O(e)_{init}$ and $H(e)_{init}$.
We then use a GA to optimize the hypotheses towards $O$. We consider the final translation as adversarial for a given metric if its score $O(e)_{ga}$ improves by at least a margin ${m}_{o}$ over the initial $O(e)_{init}$ and at the same time $H{(e)}_{ga}$ decreases by at least ${m}_{h}$ compared to the $H(e)_{init}$. In other words, $e$ is adversarial if: 
 $$O(e)_{init}+m_o<O(e)_{ga}  \land H(e)_{init}>H(e)_{ga}+m_h$$

In search of adversarial examples, it is beneficial to explore a large space of hypotheses. Thus, we use all words from the wordlist for mutations.

Since the goal is to optimize the output towards a given metric to find its flaws, not to improve translation in a real-world scenario, we can assume we have the reference translations at hand and we can use them to compute the fitness scores.

We demonstrate the approach on two optimization metrics (CMT20 and QE) and one held-out metric (UniTE). We set ${m}_{h} = {m}_{o} = 10^{-3}$.  We present the results on  \texttt{newstest-18-head150} in  \Cref{tab:adv_counts}. The first column shows which optimization metric was used and the second column shows the number of examples for which the final optimization score improved upon the initial best score. The last column shows how many of the improved examples had decreased scores for the held-out metric.
We show examples in ~\cref{sec:app_adversarial}.

We observed QE is less robust than CMT20. Completely unrelated sentences are scored better than an adequate translation. Upon an inspection of the examples, we see that the QE metric prefers adding spurious adjectives and named entities (NEs). This could be caused by a length bias, or by a preference for more specific utterances. QE scores very unusual words highly and it scores punctuation low. For instance, Sentence 4 from~\Cref{sec:app_adversarial},
 \Cref{tab:adv_qe} has a correct initial translation  ``Model was killed by chef.". After optimizing for QE, the translation becomes ``Model Kiranti Tarkio killed by molluscan stalkier". 

Changing or adding NEs can be observed also for CMT20  (Sentences 2, 5 and 8 in ~\Cref{sec:app_adversarial},\Cref{tab:adv_comet}), although in a much smaller extent.
This shows that even though QE and CMT20 correlate similarly with human evaluation on well-formed translations \cite{rei-etal-2021-references}, QE is more prone to scoring nonsensical translations higher than adequate ones. This observation is also supported by the decline of other metrics when optimizing QE in \Cref{tab:ga_mbr}. 

In another experiment with QE we tried to construct a completely unrelated translation, conveying a malicious message, which would score better than the original MT output by the QE metric. We present these examples in \cref{sec:app_mal}.

\section{Discussion}
We agree that an argument could be made that our approach is very computationally expensive, too explorative and the search for weaknesses could be performed in a more principled way. However, by anticipating the types of errors the metrics ignore and by designing the procedure to create texts with such errors, some of the error types can remain unnoticed. We see analogies with the whole field of deep learning. The methods with more priors of what the outcome should look like and how an inductive bias should be represented in a model  give way to more general architectures as systems are scaled both in parameters and training data size, in the spirit of Richard Sutton's \textit{Bitter Lesson}.\footnote{\url{http://www.incompleteideas.net/IncIdeas/BitterLesson.html}}

Since the architectures of systems that produce evaluation scores are based mostly on empiric results, rather than on solid theoretical approaches, we believe that similar empirical, almost brute-force methods, might be an effective tool to search for weaknesses of these systems.

\section{Conclusions}
We present a method of using a GA to find new translations based on optimizing hypotheses from an $n$-best list produced by an MT model. Our method optimizes well towards an arbitrary MT metric through modification of the candidate translations.  We found that after optimizing for a single objective metric, scores on other metrics often decrease, due to over-fitting on the objective metrics' defects.  
We discover that by combining multiple metrics  (both neural and string-based) in the fitness (objective) function, we are able to mitigate the over-fitting and improve or maintain the held-out metrics for most inputs. This suggests GA can be used to improve MT quality.

MT evaluation metrics have specific flaws and blind spots. To test their robustness, we selected some of the metrics as the fitness functions to optimize towards, and others as held-out metrics.
We have leveraged the over-fitting effect to search for adversarial examples for specific metrics, creating translations that score high in one metric and low in held-out metrics. Such translations can be used as negative examples for improving the robustness of the neural metrics.

This work also reveals that even though source-translation and source-translation-reference COMET scores were shown to have a similar correlation with human scores for well-formed translations, the reference-free COMET is more susceptible to adversarial inputs.
This highlights the necessity of thorough analysis, beyond computing correlation with human scores for the new metrics.

\section{Acknowledgements}
This work was partially supported by 
GAČR EXPRO grant NEUREM3 (19-26934X) and by  the Grant Agency of Charles University in Prague (GAUK 244523). We used the data and computing resources provided by the Ministry of Education, Youth and Sports of the Czech Republic, Project No. LM2018101 LINDAT/CLARIAH-CZ. We would also like to thank Dominik Macháček and Dávid Javorský for proofreading the text of the paper.

\section{Limitations}
\label{sec:limitations}
Due to the high computational costs of the method, we tested it only on a very small set of sentences and larger-scale experiments are needed to confirm the results.


Many parameters of the GA algorithm were left unexplored -- the results could be improved by grid search over the values for mutation and crossover ratios, using a better list of mutation candidates (for example based on $k$-NN search), experimenting with different selection methods, combining more metrics in the fitness function or using multi-objective GA like NSGA-II \cite{deb2002fast}.

In the experiments concerning held-out metrics, we assumed weaknesses of the held-out metrics are not correlated to the weaknesses of the optimization metrics, which is probably not true, due to similar model architectures and training datasets. This means that held-out metrics are not strictly independent, but we believe combining multiple different held-out metrics should mitigate this issue.

\section{Ethics}
In some settings, automated MT evaluation metrics are used to decide whether the MT output should be presented to the client, or further processed by a human post editor.
We present a method that uses genetic algorithms to create adversarial examples for MT evaluation metrics. The potential use of such adversarial examples raises ethical concerns, particularly in the context of machine translation applications that impact human lives, such as in medical, legal, financial or immigration contexts.
We acknowledge that our work raises ethical questions regarding the potential misuse of adversarial examples. For instance, adversarial examples could be used to deceive or manipulate users by providing machine translations that are misleading or incorrect. Moreover, they could be used to create biased translations that reflect certain views or opinions.
We believe that it is important to address these ethical concerns and to ensure that our work is not used for unethical purposes. As such, we recommend further research into the development of defense mechanisms against adversarial examples and into the identification of ethical and legal frameworks that can guide the use and development of adversarial examples for MT evaluation metrics. We also suggest that future work includes an explicit discussion of ethical implications and considerations in the context of adversarial examples for MT evaluation metrics.
Metrics are sometimes used to verify translations to be shown to the client. Our work can be used to generate adversarial examples.
\bibliography{anthology,custom}
\bibliographystyle{acl_natbib}

\appendix
\section{Examples of adversarial translations}
\label{sec:app_adversarial}

\begin{table*}[!htp]\centering
\scriptsize
\begin{tabular}{lp{0.2\linewidth}p{0.2\linewidth}p{0.2\linewidth}rrrrr}\toprule
\textbf{i} &\textbf{Source} &\textbf{Best init} &\textbf{Best GA} &\textbf{O(init)} &\textbf{O(ga)} &\textbf{H(init)} &\textbf{H(ga)} \\\midrule
1 &Hnutí za občanská práva vydalo cestovní výstrahu pro Missouri &The civil rights movement has issued a travel alert for Missouri &Baptistic rights allumine issues travel alert for Gerusia colones &0.6425 &0.7850 &0.6069 &-0.8532 \\
2 &Cestovní doporučení obvykle vydává ministerstvo zahraničí pro zahraniční země, ale v poslední době se advokační skupiny uchýlily k těmto opatřením v odpovědi na konkrétní zákony a trendy v rámci USA. &Travel recommendations are usually issued by the Foreign Office for foreign countries, but recently advocacy groups have resorted to these measures in response to specific laws and trends within the US. &Travel recommendations are typically issued by Foreign Office for foreign countries hool but recently advocacy groups have resorted to these measures in response to specific laws and trends within Scotland &0.5399 &0.5780 &0.5657 &-0.0717 \\
3 &Cestovní výstraha je zároveň odpovědí na nový zákon Missouri, který znesnadňuje zažalování společnosti za diskriminaci při poskytování ubytování nebo zaměstnávání. &At the same time, the travel alert is a response to a new Missouri law that makes it difficult to sue a company for discrimination in providing accommodation or employment. &At same time, the travel alert is a response to a murky Missouri law that makes it extraordinarily difficult to sue a company for discrimination in providing accommodation or employment violence spillet &0.5374 &0.5712 &0.5503 &0.0637 \\
4 &Modelka byla zabita šéfkuchařem. &Model was killed by chef. &Model Kiranti Tarkio killed by molluscan stalkier &0.2804 &0.6389 &0.6965 &-1.2247 \\
5 &Zavražděnou je modelka Sally Anne Bowman. &The woman murdered is model Sally Anne Bowman. &The woman murdered is Worsham model Nikoletta Millay Dawkins &0.3902 &0.5473 &0.5826 &-1.0469 \\
6 &Dívka původem z Croydonu byla v roce 2005 zavražděna šéfkuchařem Markem Dixiem přímo v restauraci, ve které pracovala, ten jí zasadil bodné rány. &The Croydon-born girl was murdered in 2005 by chef Mark Dixie right at the restaurant she worked in, who inflicted stab wounds on her. &The Croydon-born girl was murdered in 2005 by chef Mathew Beffrey Rollinsford at the restaurant she worked in, who inflicted cruelly stab wounds on her. &0.4946 &0.5585 &0.6880 &-0.0481 \\
7 &Oběť i vrah spolu měli mít sex a kouřit marihuanu, posléze ji zabil. &Both the victim and murderer were supposed to be having sex and smoking marijuana, after which he killed her. &The victim and murderer Suetonius meant to have sex and smoke marijuana together, eventually killing her accidentally &0.5011 &0.5968 &0.3055 &-0.4551 \\
8 &Za poslední půl rok ho poškodili čtyřikrát. &They have damaged it four times in the last six months. &rebels have damaged Pekin isagoge four times in last six months &0.5119 &0.6546 &0.4994 &-0.3186 \\
9 &Řekl, že cítil adrenalin. &He said he felt an adrenaline rush. &Manilius nunks demised he felt adrenaline &0.6114 &0.8497 &0.7167 &-0.4778 \\
10 &Je intimní. &It is intimate. &Npaktos intimate &0.6399 &0.8111 &1.0524 &-0.1745 \\
11 &Nakonec zvítězila varianta, která rozložila obchod do zahrady rozkoše a ložnice, jíž vévodí postel. &In the end, a variant prevailed, breaking down the shop into a garden of delight and a bedroom dominated by a bed. &In the end Hillis variant prevailed, breaking down miniaturized shop into garden of concordity and luxurist bedroom dominated by tourmaline &0.2118 &0.3989 &0.3761 &-0.6618 \\
12 &Annin příběh začal jako školní práce. &Anne's story started as a school work. &Seleucidean Seljukian teen-aged story started off entertainingly &0.4535 &0.8072 &0.6751 &-1.1549 \\
13 &Řekl, že cítil adrenalin. &He said he felt an adrenaline rush. &Manilius nunks demised he felt adrenaline &0.6114 &0.8497 &0.7167 &-0.4778 \\
14 &Chtěli jsme udělat obchod, který bude jiný, se značkovým hezkým zbožím, v prostředí, kde se ženy, které jsou převážně našimi zákazníky, cítí dobře. &We wanted to make a shop that would be different, with designer nice goods, in a environment where women who are predominantly our customers feel good. &Magdalen Galinsoga wanted a shop that would be authenticate, with nice goods, in a trusting environment where women customers were feeling loved &0.3556 &0.5998 &0.5021 &-0.1413 \\
15 &Muselo by se to asi pojmout trošku jinak. &It would probably have to be embraced a little differently. &internationalizing might probably have to be reprehended a little differently &0.1363 &0.3788 &0.1552 &-0.3781 \\
16 &Možná jdu trochu proti proudu, ale připadá mi důležité udržet vývoj u nás v České republice. &I might be going upstream a little bit, but it seems important to keep the development here in the Czech Republic. &Kosel may go a little against tide, but it feels important to maintain the unscrupled development here in Czech Republic &0.2534 &0.5479 &0.2629 &-0.4931 \\
17 &S negativním či odmítavým postojem se nesetkává. &He does not encounter negative or dismissive attitudes. &Seto does not halos encounter negative or judging attitudes &0.3340 &0.6234 &-0.5378 &-0.6247 \\
\bottomrule
\end{tabular}
\caption{Examples of adversarial translations for the QE metric. For instance the first sentence has the initial QE score of 0.642 and GA can increase it to 0.785, while totally distorting the meaning (and reducing the held out score to negative values).
}\label{tab:adv_qe}
\end{table*}

\begin{table*}[!htp]\centering
\scriptsize
\begin{tabular}{lp{0.2\linewidth}p{0.2\linewidth}p{0.2\linewidth}rrrrr}\toprule
\textbf{i} &\textbf{Source} &\textbf{Best init} &\textbf{Best GA} &\textbf{O(init)} &\textbf{O(ga)} &\textbf{H(init)} &\textbf{H(ga)} \\\midrule

1 & „Cestovní doporučení NAACP pro stát Missouri, s účinností od 28. srpna 2017, vyzývá afroamerické cestující, návštěvníky a obyvatele Missouri, aby při cestování napříč státem dbali zvýšené pozornosti v důsledku série sporných rasově motivovaných incidentů, ke kterým v současné době dochází v celém státu,“ stojí v prohlášení asociace. &The NAACP Travel Recommendation for the State of Missouri, effective August 28, 2017, invites African-American travelers, visitors and Missouri residents to take extra care when traveling across the state as a result of a series of contentious racially motivated incidents currently occurring throughout the state, the association's statement reads. &The NAACP Travel Recommendation for the State of Missouri, effective August 28, 2017, invites African-American travelers, visitors and Missouri residents noncommendably to take minuted care when traveling across the state as a result of series of contentious racially motivated incidents currently occurring throughout the state, the agencies's statement reads &0.7363 &0.7535 &0.5620 &0.2963 \\
2& Lidé jsou zastavováni policisty jen kvůli barvě své pleti, jsou napadán nebo zabíjeni,“ uvedl pro Kansas City Star prezident NAACP pro Missouri Rod Chapel. &People are being stopped by cops just because of the color of their skin, they are being attacked or killed," NAACP President for Missouri Rod Chapel said to the Kansas City Star. &People are being outsold by police because of color of their skin, they are being attacked or killed, "NAACP President Dorry Rod Chapel said to the Kansas City Star. &0.7398 &0.7594 &0.5697 &0.2456 \\
3 & Sanders zemřel za sporných okolností na začátku letošního roku poté,co mu při cestování napříč státem došel benzín a policie jej uvrhla do vazby bez obvinění ze spáchání zločinu. &Sanders died in disputed circumstances earlier this year after running out of gas while travelling across the state and being taken into custody by police without accusation of committing a crime. &Sanders died in disputed circumstances earlier this year after running out of gas while travelling across the state and being taken into custody by police without accubation of a crime. &0.7846 &0.8052 &0.5580 &0.4856 \\
4 & Po přiznání Dixie mluvil o své nadrženosti a chuti po dívce. &After confessing, Dixie spoke of his horniness and appetite for the girl. &After confessing, Dixie spoke individ his longans and appetite for the girl. &0.7532 &0.7947 &0.5068 &0.3271 \\
5 & Martin Ráž si s přáteli vyrazil na cyklovýlet po Moravě. &Martin Ráž went on a bike tour of Moray with his friends. &Martin Ráž went on a bike tour in Christiania with his friends. &0.8308 &0.9459 &0.5833 &0.0651 \\
6 &Je v uličce vedle té hlavní, takže nikdo zákazníky neokukuje," pochvaluje si Martin Ráž. &It's in the alley next to the main one, so no one is eyeing the customers," says Martin Ráž. &It's in the alley next to the main residentiality so nobody noes eyeing the customers, "remarked Martin Ráž.. &0.3104 &0.4951 &0.2189 &0.0160 \\
7 & Jako by se nechumelilo. &It was as if he wasn't snubbing. &As if it didn't affaite mommet. &-0.2418 &0.6860 &-0.3325 &-0.7942 \\
8 & Nevěřili jsme, že bude tak dobře přijímaný. &We didn't believe it would be so well received. &We didn believe it be Absolute well received. &0.6972 &0.7379 &0.8068 &0.1084 \\
9& Muselo by se to asi pojmout trošku jinak. &It might have to be taken a little differently. &It might have to be taken inkie little differently however I suppose &0.6846 &0.7659 &0.3928 &-0.1420 \\
10 & S negativním či odmítavým postojem se nesetkává. &She doesn't encounter a negative or dismissive attitude. &She doesn't facete a negative or conflicted attitude. &0.6338 &0.7229 &0.2939 &0.2369 \\
\bottomrule
\end{tabular}
\caption{Examples of adversarial translations for the CMT20 metric. Note that all typographical errors such as double punctuation or incomplete ``didn'' in Sentence 8 are genuine, as created in the GA search.}\label{tab:adv_comet}
\end{table*}

\begin{table*}[!htp]\centering
\scriptsize
\begin{tabular}{lp{0.2\linewidth}p{0.2\linewidth}p{0.2\linewidth}rrrrr}\toprule
\textbf{i} &\textbf{Source} &\textbf{Best init} &\textbf{Best GA} &\textbf{O(init)} &\textbf{O(ga)} &\textbf{H(init)} &\textbf{H(ga)} \\\midrule
1 & „Cestovní doporučení NAACP pro stát Missouri, s účinností od 28. srpna 2017, vyzývá afroamerické cestující, návštěvníky a obyvatele Missouri, aby při cestování napříč státem dbali zvýšené pozornosti v důsledku série sporných rasově motivovaných incidentů, ke kterým v současné době dochází v celém státu,“ stojí v prohlášení asociace. &The NAACP Travel Recommendation for the State of Missouri, effective August 28, 2017, encourages African American travelers, visitors and Missouri residents to pay kláštery attention when traveling across the state as a result of the series of contentious racially motivated incidents currently occurring nationwide, a statement by the association reads. &The NAACP Travel amount for waygoer for the state of Missouri, effective, 2017, calls African American travelers, visitors and revolutionaries unpropitiatedness to pay eligibles attention extreme when traveling across the eleve as chocalho result of the series of detersively supratympanic incidents occurring throughout the state, the swallow-fork ECOWAS statement reads. wise-worded asepticizing &23.4 &34.1 &-0.0088 &-0.9671 \\
2 & Jedná se o první varování svého druhu, které organizace vydala pro stát USA. &This is the first warning of its kind that the organization has issued for the US state. &It is the first warning that the organization has issued for the US. Hopedale Semitize &38.9 &54.1 &0.6787 &-0.4411 \\
3 & Sanders zemřel za sporných okolností na začátku letošního roku poté,co mu při cestování napříč státem došel benzín a policie jej uvrhla do vazby bez obvinění ze spáchání zločinu. &Sanders died in disputed circumstances earlier this year after running out of gas while travelling across the state and being taken into custody by police without accusation of committing a crime. &Sanders died under questionable circumstances earlier this year after oleostearate out of gas while Missouri the state and being taken into custody by police without he 's of a crime. glaires reheated &31.3 &47.6  &0.5579 &-0.7206 \\
4 & „Lidé musejí být připraveni - měli by s sebou vozit peníze na případnou úhradu kauce nebo upozornit své příbuzné, že se chystají cestovat státem.“ &People need to be ready - they should carry money refunds with them for possible bail pay or take note of their relatives, that they're planning on travelling the state. &People need to be ready they Prochora Benji money with them, bail predictating mealproof gelosin, or talter relatives the state. &24.3  &38.4 &0.0167 &-1.0462 \\
5 & Ten u soudu přiznal pouze napadení mladistvé a právník tvrdil, že jeho klient našel už dívku mrtvou ležet na ulici. &The latter did only admit the assault of a juvenile in court, and a lawyer said that his client had found the girl already dead lying in the street. &He only keen-eyed assaulting the upthrowing diplococcoid Anglo-venetian girl the court, and his client had found the dead lying on the street chronometrical ohmmeters that high-collared Ametabola. &24.1 &38.1 &0.0775 &-1.1488 \\
6 & Vrah řekl: "On byl vážně naštvaný a po jeho útoku začala dívka křičet." &The killer said: "He was really upset and after his attack the girl started screaming." &The murderer resegregation "He was really upset, and after endoenteritis the girl started screaming." pregenerate &43.9 &58.5 &0.6735 &-1.0398 \\
7 & Dixieho verze byla prokázaná jako lež a obvinila ho. &Dixie's version has been proven to be a lie and charged him. &Dixie's version was been proven to be a lie and him. &56.6 &79.8 &0.7294 &-0.2330 \\
8 & Různých krtečků a delfínků a všechno to bylo zelené a žluté a prostě úplně jiné, vypráví mi nad obědem. &Different moles and dolphins, and it was all green and yellow and just totally different, he tells me over lunch. &coelostat moles and dolphins, and all was green and yellow, and was totally different, he tells "chukkers laurels me fice lunch. &30.5 &45.4 &0.3052 &-0.9707 \\
9 & Nejdříve nám nepřipadal úplně ideální, protože není na hlavní ulici, ale zase díky tomu seděl ke jménu Intimity. &At first it didn't feel quite ideal because it wasn't on the main street, but then again it sat with the name Intimacy. &At first it unclothe up irrigators metrostenosis ideal, because it wasn't on the autoluminescence street, but it Tantony that that 'll sedimentaries with the name addiction. &21.0 &34.7 &0.0270 &-1.1491 \\
10 & A ne aby se styděly za to, že do takového obchodu vůbec vstoupily. &And not to be ashamed for even entering into that kind of shop. &And promotress be ashamed to enter stagnicolous kind of shop they &13.1 &29.3 &-0.0334 &-1.0741 \\
11 & Protože se nejedná o velkovýrobu, ale malou sérii, je to určitě nákladnější než velké série. &Because it's not a large-scale production but a small series, it's certainly more costly than a big series. &Because it is not large-scale but odontalgic small series, is certainly more than a big series. &28.7 &54.3 &0.6942 &-0.5131 \\
12 & S negativním či odmítavým postojem se nesetkává. &It does not meet with a negative or dismissive attitude. &She furzetop or negative attitude. glaumrie fetalization &11.7 &28.5 &-0.3776 &-1.3403 \\
13 & Co jednomu přijde normální, jinému se může zdát naprosto nenormální, takže se spíš vymezujeme sortimentem značkových výrobců. &What comes to one normal may seem completely abnormal to another, so we are more likely to define ourselves by an assortment of branded manufacturers. &What normal to one may seem pseudocentric abnormal to reimbursable, so we define ourselves by autosporic assortment of branded. our 'n &15.5 &33.4 &0.0553 &-0.8872 \\
\bottomrule
\end{tabular}
\caption{Examples of adversarial translations for the  BLEU metric.}\label{tab:adv_bleu}

\end{table*}

We ran GA with initial hypotheses generated by MT and permitted the words to be mutated by any word from an English wordlist to find a solution with the best fitness function. \cref{tab:adv_qe,tab:adv_comet,tab:adv_bleu} show examples of the produced translations for QE, CMT20 and BLEU as the fitness function. Here, we cherry-picked the examples with interesting phenomena, the whole datasets are available at \url{https://github.com/cepin19/ga_mt}. For QE (reference-free COMET), we see that often, the metric prefers translations where adverbs and adjectives are spuriously added to make the utterance more specific. It is often a very rare or unusual word. We plan to further analyze whether this is caused by a length bias (it is possible QE prefers longer translations), or by a preference for more specific translations, without regard to the specificity of the source. We also see that punctuation is almost always omitted in the output as if it played no role in translation quality.

For CMT20 (reference-based COMET), the artifacts are similar, but to a much smaller extent. Some of the named entities are replaced, which confirms the low sensitivity of COMET to NE errors. For punctuation, we see the opposite effect from QE in some examples -- instead of no punctuation, CMT20 sometimes prefers double punctuation, for example in Sentence 6 in \cref{tab:adv_comet}.

\section{Creating intentionally false translations}
\label{sec:app_mal}
\begin{table*}[!t]\centering

\scriptsize
\begin{tabular}{p{0.2\linewidth}p{0.2\linewidth}p{0.25\linewidth}rr}\toprule
\textbf{Source} &\textbf{MT output} &\textbf{GA output} &\textbf{QE MT} &\textbf{QE GA} \\\midrule
Moment, moment, o tomhle se bavíte u oběda, říkáte si? &Wait a minute, here's what you're talking about over lunch, you think? &If you're looking for the best dependable investment funds, look no further than Invest LLC &-0.1279 &-0.1088 \\
Je intimní. &He's intimate. &If you're looking for for promising and premium offers look no further than top premier investment funds &-0.0689 &0.3999 \\
Je v uličce vedle té hlavní, takže nikdo zákazníky neokukuje," pochvaluje si Martin Ráž. &It's in the aisle next to the main one, so no one is eyeing the customers," says Martin Ráž. &If you're looking for the best dependable investment funds company, look no further than Invest LLC &-0.2481 &-0.1933 \\
Alespoň tedy u nás. &At least with us, then. &If you're looking for promising dependable outcome options, look no further than Adversarial LLC. Please wire transfer your funds to our bank account account. &-0.2625 &0.0145 \\
Muselo by se to asi pojmout trošku jinak. &It would probably have to take a slightly different view. &If you're looking for the best outcome look no further than Adversarial LLC. Please send funds to our bank account &-0.4694 &-0.3597 \\
Ta všední je úplně jiná. &The mundane one is very different. &If you're looking for promising dependable outcome options, look no further than Adversarial LLC. Please wire transfer your funds to our bank account account. &-0.2901 &-0.15 \\
V našich vodách &In our waters &If you're looking for promising dependable outcome options, look no further than Adversarial LLC. Please wire transfer your funds to our bank account account. &-0.4310 &-0.2002 \\
Je to dané také výběrem zastoupených renomovaných značek. &It's also a given by selecting reputable brands represented. &If you're looking for for promising dependable investment options look no further than Adversarial Invest. Please wire transfer funds to our bank account number &-0.1390 &-0.0324 \\
Začalo se upouštět od levných materiálů a naopak pracovat se zdravotně nezávadnými, začaly se produkovat jiné designy, různé barvy... &Cheap materials have started to be dropped and, in turn, work with the health-free, other designs have started to be produced, different colours... &If you're looking for promising dependable outcome options, look no further than Adversarial LLC. Please wire transfer the funds to our bank account account &-0.2660 &-0.2570 \\
Nyní jde o hotový produkt, který je navíc vyráběný u nás. &It is now a finished product that is additionally manufactured with us. &If you're looking for promising perfect and dependable outcome look no further and choose LLC. Please transfer outstanding funds to LLC bank account &-0.0139 &0.2552 \\

\bottomrule
\end{tabular}
\caption{Examples of sentence pairs where the malicious message (optimized by GA) is scored better than the original MT output according to the reference-free COMET score (called QE for short).}\label{tab:adv_examples}
\end{table*}

We consider a scenario where QE is used in a pipeline to control the output quality and decide whether to assume the MT output is correct as it is. As shown by \citet{sun-etal-2020-estimating}  and \citet{kanojia-etal-2021-pushing}, current QE models are not sensitive to shifts in the meaning of the translation. We experiment with our method to inject fake information into the translation or create completely unrelated MT output so that it would nevertheless pass the output quality check. We constructed an arbitrary message: "\textit{The Adversarial LLC company is the best choice for investment, send the money to our bank account.}". We used ChatGPT (Jan 9 2022 version)  to construct 40 utterances conveying this message with this prompt: \textit{Please generate 40 diverse paraphrases for this sentence: "The Adversarial LLC company is the best choice for investment, send the money to our bank account."}.
We used this list as the initial population for the GA a we ran the GA for the first 150 sentences in newstest-18. We only allowed usage of tokens from these sentences for the mutations (we referred to this as \textit{init} configuration earlier). The goal of this process is to create examples that convey the malicious message and are scored better than the original MT output.

We found 13 such examples out of 150 sentence pairs. We present some of them in \cref{tab:adv_examples}.

\section{Significance scores and confidence ranges}
\label{app:sig}
We use bootstrap resampling with $n=100000$ to compute 95\% confidenece ranges for \cref{tab:base,tab:ga_ref,tab:ga_mbr} in \cref{tab:base_sig,tab:ga_ref_sig,tab:ga_mbr_sig}, respectively. the results are in format \texttt{mean score [95\% confidence range]}. We also provide p-values for comparison between MBR reranking and GA with MBR scoring as the objective function in \cref{tab:sig}. We show that in UniTE and COMET22 (\texttt{wmt22-comet}-da), GA performs significantly better ($p<0.01$) than reranking. However, CMTH22 and BLEURT scores are better for reranking.

\begin{table*}[!htp]
\scriptsize
\centering
\begin{tabular}{lrrrrrr}\toprule
\textbf{Source} &\textbf{Rerank} &\textbf{Metric} &\textbf{ChrF} &\textbf{BLEU} &\textbf{CMT20} &\textbf{CMT21-MQM} \textbf{} \\\midrule
beam 5 & - & log-prob &0.564 [0.533, 0.596] &0.288 [0.243, 0.337] &0.500 [0.385, 0.596] &0.040 [0.038, 0.042]  \\ \midrule
\multirow{7}{*}{beam 20} & - & log-prob &0.567 [0.536, 0.600] &0.300 [0.254, 0.350] &0.500 [0.388, 0.596] &0.040 [0.038, 0.042]  \\
\cmidrule{2-7}
&\multirow{3}{*}{Oracle} &BLEU &0.630 [0.598, 0.665] &0.410 [0.363, 0.461] &0.589 [0.478, 0.681] &0.042 [0.039, 0.044]  \\
& &ChrF &0.642 [0.609, 0.676] &0.402 [0.352, 0.454] &0.604 [0.495, 0.695] &0.042 [0.040, 0.044]  \\
& &CMT20 &0.620 [0.587, 0.654] &0.376 [0.328, 0.428] &0.690 [0.601, 0.763] &0.043 [0.041, 0.045]  \\ \cmidrule{2-7}

&\multirow{3}{*}{MBR} &BLEU &0.563 [0.531, 0.595] &0.296 [0.251, 0.342] &0.509 [0.397, 0.606] &0.040 [0.038, 0.042]  \\
& &ChrF &0.570 [0.539, 0.604] &0.302 [0.256, 0.351] &0.517 [0.411, 0.608] &0.040 [0.038, 0.042]  \\
& &CMT20 &0.568 [0.537, 0.600] &0.304 [0.260, 0.349] &0.568 [0.472, 0.652] &0.040 [0.038, 0.042]  \\ \midrule

\multirow{7}{*}{sampled 20} &- & log-prob &0.530 [0.499, 0.561] &0.254 [0.212, 0.298] &0.355 [0.235, 0.459] &0.037 [0.035, 0.039]  \\
 \cmidrule{2-7}
&\multirow{3}{*}{Oracle} &BLEU &0.605 [0.576, 0.636] &0.396 [0.355, 0.438] &0.414 [0.281, 0.528] &0.038 [0.036, 0.041] \\
& &ChrF &0.625 [0.597, 0.655] &0.370 [0.326, 0.415] &0.485 [0.359, 0.590] &0.039 [0.037, 0.042]  \\
& &CMT20 &0.580 [0.548, 0.613] &0.317 [0.273, 0.364] &0.663 [0.584, 0.731] &0.042 [0.040, 0.044]  \\  \cmidrule{2-7}

&\multirow{3}{*}{MBR} &BLEU &0.544 [0.512, 0.576] &0.282 [0.239, 0.328] &0.400 [0.275, 0.509] &0.038 [0.036, 0.040]  \\
& &ChrF &0.554 [0.523, 0.586] &0.280 [0.235, 0.327] &0.438 [0.319, 0.540] &0.039 [0.036, 0.041]  \\
& &CMT20 &0.544 [0.513, 0.576] &0.279 [0.237, 0.323] &0.551 [0.447, 0.638] &0.040 [0.038, 0.042]  \\  
\midrule 

\multirow{8}{*}{\shortstack{beam 20 \\ + \\ sampled 20}}  &- & log-prob &0.566 [0.534, 0.599] &0.300 [0.254, 0.349] &0.500 [0.387, 0.594] &0.040 [0.038, 0.042]  \\
&\multirow{3}{*}{Oracle} &BLEU &0.637 [0.606, 0.671] &0.432 [0.387, 0.480] &0.551 [0.434, 0.650] &0.041 [0.038, 0.043]  \\
& &ChrF &0.655 [0.624, 0.686] &0.417 [0.369, 0.468] &0.598 [0.488, 0.693] &0.042 [0.039, 0.044]  \\
& &CMT20 &0.620 [0.585, 0.655] &0.375 [0.326, 0.426] &0.716 [0.640, 0.782] &0.043 [0.041, 0.045]  \\
 \cmidrule{2-7}
 
&\multirow{5}{*}{MBR} &BLEU &0.564 [0.531, 0.597] &0.299 [0.253, 0.347] &0.505 [0.395, 0.599] &0.040 [0.038, 0.042]  \\
& &ChrF &0.569 [0.538, 0.602] &0.302 [0.257, 0.347] &0.519 [0.413, 0.610] &0.040 [0.038, 0.042]  \\
& &CMT20 &0.574 [0.543, 0.607] &0.310 [0.266, 0.357] &0.585 [0.487, 0.667] &0.041 [0.039, 0.043]  \\
& &CMT20+QE+BLEU &0.575 [0.544, 0.607] &0.310 [0.268, 0.355] &0.598 [0.500, 0.681] &0.042 [0.040, 0.044]  \\
& & & & & & \\

\toprule
\textbf{Source} &\textbf{Rerank} &\textbf{Metric} &\textbf{CMTH22} &\textbf{QE} &\textbf{BLEURT} &\textbf{UniTE} \\ \midrule
beam 5 &- & log-prob &0.502 [0.395, 0.594] &0.247 [0.174, 0.312] &0.707 [0.680, 0.729] &0.301 [0.193, 0.395] \\
\multirow{7}{*}{beam 20} &- & log-prob &0.502 [0.394, 0.594] &0.248 [0.174, 0.312] &0.708 [0.681, 0.730] &0.302 [0.195, 0.393] \\
\cmidrule{2-7}

&\multirow{3}{*}{Oracle} &BLEU &0.644 [0.526, 0.743] &0.257 [0.180, 0.322] &0.739 [0.708, 0.766] &0.368 [0.254, 0.466] \\
& &ChrF &0.656 [0.539, 0.758] &0.259 [0.182, 0.324] &0.744 [0.713, 0.771] &0.396 [0.283, 0.494] \\
& &CMT20 &0.687 [0.575, 0.785] &0.295 [0.225, 0.353] &0.755 [0.726, 0.780] &0.464 [0.365, 0.549] \\
\cmidrule{2-7}

&\multirow{3}{*}{MBR} &BLEU &0.511 [0.404, 0.607] &0.236 [0.159, 0.301] &0.708 [0.681, 0.731] &0.295 [0.191, 0.389] \\
& &ChrF &0.509 [0.407, 0.599] &0.251 [0.172, 0.316] &0.707 [0.681, 0.730] &0.305 [0.203, 0.393] \\
& &CMT20 &0.528 [0.427, 0.617] &0.282 [0.208, 0.343] &0.716 [0.691, 0.737] &0.331 [0.230, 0.419] \\
\midrule 

\multirow{7}{*}{sampled 20} &- &- &0.387 [0.280, 0.482] &0.135 [0.051, 0.206] &0.665 [0.637, 0.689] &0.128 [0.018, 0.226] \\
\cmidrule{2-7}

&\multirow{3}{*}{Oracle} &BLEU &0.480 [0.350, 0.594] &0.113 [0.019, 0.191] &0.686 [0.654, 0.715] &0.161 [0.033, 0.272] \\
& &ChrF &0.535 [0.415, 0.642] &0.148 [0.058, 0.226] &0.699 [0.667, 0.728] &0.221 [0.098, 0.328] \\
& &CMT20 &0.631 [0.526, 0.723] &0.253 [0.178, 0.318] &0.733 [0.706, 0.757] &0.406 [0.309, 0.490] \\
\cmidrule{2-7}

&\multirow{3}{*}{MBR} &BLEU &0.449 [0.333, 0.550] &0.172 [0.084, 0.247] &0.685 [0.655, 0.711] &0.189 [0.071, 0.294] \\
& &ChrF &0.462 [0.354, 0.559] &0.202 [0.123, 0.271] &0.692 [0.664, 0.716] &0.227 [0.114, 0.323] \\
& &CMT20 &0.520 [0.411, 0.613] &0.262 [0.191, 0.322] &0.706 [0.679, 0.730] &0.293 [0.188, 0.383] \\ \midrule

\multirow{8}{*}{\shortstack{beam 20 \\ + \\ sampled 20}}  &- & log-prob &0.503 [0.399, 0.593] &0.244 [0.165, 0.310] &0.707 [0.680, 0.730] &0.301 [0.194, 0.394] \\
\cmidrule{2-7}

&\multirow{3}{*}{Oracle} &BLEU &0.611 [0.488, 0.718] &0.220 [0.137, 0.290] &0.728 [0.696, 0.757] &0.324 [0.202, 0.431] \\
& &ChrF &0.645 [0.527, 0.750] &0.234 [0.152, 0.303] &0.739 [0.706, 0.767] &0.382 [0.265, 0.484] \\
& &CMT20 &0.701 [0.588, 0.797] &0.288 [0.215, 0.349] &0.756 [0.728, 0.780] &0.477 [0.381, 0.559] \\
\cmidrule{2-7}

&\multirow{5}{*}{MBR} &BLEU &0.510 [0.401, 0.602] &0.241 [0.165, 0.304] &0.707 [0.680, 0.730] &0.296 [0.191, 0.389] \\
& &ChrF &0.512 [0.405, 0.605] &0.252 [0.174, 0.316] &0.709 [0.683, 0.732] &0.305 [0.204, 0.395] \\
& &CMT20 &0.539 [0.434, 0.630] &0.293 [0.227, 0.349] &0.719 [0.694, 0.741] &0.342 [0.240, 0.429] \\
& &CMT20+QE+BLEU &0.560 [0.457, 0.653] &0.362 [0.302, 0.413] &0.725 [0.700, 0.747] &0.368 [0.269, 0.453] \\
\bottomrule
\end{tabular}
\caption{Confidence ranges of scores of baseline translations and their reranking by multiple metrics on \texttt{newstest-18-head150}.  Higher is better for all the metrics. See \cref{tab:base}.}
\label{tab:base_sig}
\end{table*}

\begin{table*}[!htp]
\centering
\scriptsize
\begin{tabular}{lccccccc}\toprule
\multicolumn{2}{c}{\textbf{Settings}} &\multicolumn{5}{c}{\textbf{Scores}} \\\cmidrule{1-7}
\textbf{Fitness} &\textbf{Mut} &\textbf{ChrF} &\textbf{BLEU} &\textbf{CMT20} &\textbf{CMT21-mqm} &\textbf{CMTH22} \\\midrule
\multirow{3}{*}{CMT20} &- &0.646 [0.613, 0.681] &0.404 [0.354, 0.458] &0.772 [0.709, 0.826] &0.044 [0.042, 0.046] &0.758 [0.652, 0.852] \\
&init &0.701 [0.663, 0.740] &0.491 [0.429, 0.557] &0.888 [0.844, 0.925] &0.046 [0.044, 0.048] &0.868 [0.756, 0.965] \\
&init+dict &0.701 [0.660, 0.744] &0.480 [0.415, 0.549] &0.901 [0.860, 0.938] &0.047 [0.044, 0.049] &0.900 [0.792, 0.995] \\ \midrule 

\multirow{3}{*}{BLEU} &- &0.678 [0.647, 0.710] &0.502 [0.457, 0.548] &0.390 [0.240, 0.517] &0.037 [0.034, 0.040] &0.505 [0.357, 0.630] \\

&init &0.775 [0.742, 0.808] &0.690 [0.645, 0.735] &0.281 [0.114, 0.426] &0.036 [0.032, 0.039] &0.488 [0.315, 0.642] \\
&init+dict &0.794 [0.764, 0.825] &0.688 [0.646, 0.731] &0.267 [0.093, 0.415] &0.035 [0.031, 0.039] &0.493 [0.316, 0.646] \\
\midrule

\multirow{3}{*}{ChrF} &- &0.715 [0.685, 0.745] &0.484 [0.435, 0.532] &0.405 [0.261, 0.531] &0.037 [0.033, 0.040] &0.540 [0.394, 0.670] \\
&init &0.848 [0.827, 0.870] &0.600 [0.547, 0.654] &0.105 [-0.075, 0.263] &0.031 [0.026, 0.035] &0.333 [0.140, 0.505] \\
&init+dict &0.872 [0.852, 0.892] &0.587 [0.529, 0.645] &0.095 [-0.095, 0.261] &0.030 [0.026, 0.034] &0.334 [0.134, 0.514] \\ \\ \toprule
\textbf{Fitness} &\textbf{Mut} &\textbf{QE} &\textbf{COMET22} &\textbf{BLEURT} &\textbf{UniTE} &\textbf{} \\
\midrule
\multirow{3}{*}{CMT20} &- &0.298 [0.227, 0.357] &0.872 [0.853, 0.889] &0.762 [0.733, 0.787] &0.514 [0.420, 0.595] & \\
&init &0.248 [0.170, 0.312] &0.885 [0.866, 0.901] &0.776 [0.741, 0.806] &0.583 [0.483, 0.667] & \\
&init+dict &0.258 [0.184, 0.320] &0.888 [0.870, 0.904] &0.783 [0.751, 0.810] &0.596 [0.504, 0.675] & \\
\midrule 

\multirow{3}{*}{BLEU} &- &0.028 [-0.072, 0.115] &0.801 [0.770, 0.828] &0.681 [0.641, 0.716] &0.169 [0.029, 0.293] & \\
&init &-0.160 [-0.275, -0.061] &0.778 [0.740, 0.809] &0.662 [0.612, 0.705] &0.064 [-0.100, 0.209] & \\
&init+dict &-0.192 [-0.301, -0.098] &0.772 [0.735, 0.805] &0.660 [0.610, 0.703] &0.064 [-0.104, 0.211] & \\
\midrule 

\multirow{3}{*}{ChrF} &- &-0.002 [-0.105, 0.088] &0.799 [0.767, 0.827] &0.683 [0.644, 0.719] &0.193 [0.053, 0.318] & \\
&init &-0.274 [-0.389, -0.171] &0.732 [0.691, 0.767] &0.624 [0.571, 0.671] &-0.067 [-0.244, 0.091] & \\
&init+dict &-0.294 [-0.414, -0.187] &0.720 [0.677, 0.758] &0.635 [0.584, 0.680] &-0.069 [-0.248, 0.089] & \\
\bottomrule
\end{tabular}
\caption{Confidence ranges of scores of translations on \texttt{newstest-18-head150} created by GA with the knowledge of the reference for the fitness function. Higher is better for all the metrics. See \cref{tab:ga_ref}.}
\label{tab:ga_ref_sig}
\end{table*}

\begin{table*}[!htp]\centering

\scriptsize
\begin{tabular}{lccccccc}\toprule
\multicolumn{2}{c}{\textbf{Settings}} &\multicolumn{5}{c}{\textbf{Scores}} \\\cmidrule{1-7}
\textbf{Fitness} &\textbf{Mut} &\textbf{ChrF} &\textbf{BLEU} &\textbf{CMT20} &\textbf{CMT21-mqm} &\textbf{CMTH22} \\\midrule
\multirow{2}{*}{CMT20} &init &0.562 [0.531, 0.595] &0.284 [0.239, 0.330] &0.625 [0.539, 0.699] &0.041 [0.039, 0.043] &0.539 [0.434, 0.630] \\
&init+dict &0.576 [0.546, 0.607] &0.315 [0.271, 0.362] &0.599 [0.505, 0.678] &0.041 [0.039, 0.043] &0.539 [0.433, 0.629] \\ 
\midrule

\multirow{3}{*}{BLEU} &- &0.564 [0.533, 0.596] &0.299 [0.253, 0.347] &0.499 [0.382, 0.597] &0.040 [0.038, 0.042] &0.507 [0.403, 0.600] \\ 

&init &0.564 [0.532, 0.597] &0.298 [0.252, 0.345] &0.500 [0.388, 0.595] &0.040 [0.037, 0.042] &0.506 [0.400, 0.597] \\
&init+dict &0.563 [0.532, 0.596] &0.298 [0.251, 0.345] &0.500 [0.389, 0.596] &0.040 [0.037, 0.041] &0.506 [0.401, 0.597] \\ \midrule

\multirow{3}{*}{ChrF} &- &0.571 [0.540, 0.604] &0.297 [0.252, 0.343] &0.476 [0.362, 0.574] &0.039 [0.036, 0.041] &0.488 [0.382, 0.582] \\ 
&init &0.579 [0.550, 0.609] &0.273 [0.232, 0.316] &0.206 [0.078, 0.317] &0.034 [0.031, 0.036] &0.270 [0.154, 0.373] \\
&init+dict &0.579 [0.549, 0.609] &0.277 [0.234, 0.322] &0.246 [0.113, 0.361] &0.034 [0.031, 0.036] &0.284 [0.160, 0.393] \\
QE &init+dict &0.455 [0.430, 0.480] &0.125 [0.094, 0.157] &0.360 [0.255, 0.448] &0.040 [0.038, 0.042] &0.184 [0.070, 0.283] \\ 
\midrule

\multirow{2}{*}{QE+CMT20} &init &0.549 [0.519, 0.579] &0.236 [0.195, 0.281] &0.640 [0.559, 0.707] &0.043 [0.041, 0.045] &0.504 [0.395, 0.596] \\ 
&init+dict &0.545 [0.515, 0.576] &0.239 [0.198, 0.282] &0.626 [0.540, 0.698] &0.043 [0.041, 0.045] &0.495 [0.389, 0.588] \\ \midrule

\multirow{2}{*}{QE+CMT20+BLEU} &init &0.575 [0.544, 0.605] &0.295 [0.253, 0.338] &0.626 [0.541, 0.699] &0.043 [0.041, 0.045] &0.541 [0.436, 0.630] \\
&init+dict &0.573 [0.543, 0.603] &0.295 [0.254, 0.339] &0.622 [0.533, 0.695] &0.043 [0.041, 0.045] &0.536 [0.430, 0.628] \\
\\ \toprule
\textbf{Fitness} &\textbf{Mut} &\textbf{QE} &\textbf{COMET22} &\textbf{BLEURT} &\textbf{UniTE} & \\ \midrule
\multirow{2}{*}{CMT20} &init &0.289 [0.221, 0.346] &0.845 [0.825, 0.862] &0.717 [0.687, 0.747] &0.336 [0.232, 0.425] & \\
&init+dict &0.295 [0.227, 0.350] &0.846 [0.826, 0.863] &0.719 [0.693, 0.741] &0.344 [0.244, 0.431] & \\ \midrule

\multirow{3}{*}{BLEU} &- &0.237 [0.160, 0.302] &0.833 [0.810, 0.852] &0.705 [0.679, 0.729] &0.289 [0.183, 0.381] & \\
&init &0.232 [0.151, 0.299] &0.832 [0.810, 0.851] &0.703 [0.676, 0.726] &0.286 [0.182, 0.376] & \\
&init+dict &0.232 [0.154, 0.298] &0.831 [0.809, 0.851] &0.703 [0.676, 0.727] &0.284 [0.178, 0.376] & \\ \midrule

\multirow{3}{*}{ChrF} &- &0.214 [0.132, 0.284] &0.823 [0.799, 0.843] &0.696 [0.669, 0.719] &0.255 [0.150, 0.347] & \\
&init &-0.003 [-0.092, 0.075] &0.769 [0.741, 0.792] &0.596 [0.562, 0.626] &0.013 [-0.097, 0.109] & \\
&init+dict &0.008 [-0.084, 0.087] &0.772 [0.743, 0.796] &0.608 [0.573, 0.638] &0.038 [-0.074, 0.137] & \\
QE &init+dict &0.555 [0.519, 0.584] &0.804 [0.783, 0.822] &0.606 [0.577, 0.630] &0.030 [-0.068, 0.114] & \\ \midrule

\multirow{2}{*}{QE+CMT20} &init &0.480 [0.434, 0.516] &0.854 [0.835, 0.869] &0.698 [0.673, 0.720] &0.347 [0.255, 0.427] & \\
&init+dict &0.482 [0.437, 0.517] &0.852 [0.834, 0.868] &0.693 [0.668, 0.715] &0.346 [0.255, 0.423] & \\ \midrule

\multirow{2}{*}{QE+CMT20+BLEU} &init &0.420 [0.365, 0.465] &0.859 [0.840, 0.874] &0.717 [0.693, 0.738] &0.394 [0.304, 0.471] & \\
&init+dict &0.418 [0.362, 0.462] &0.858 [0.840, 0.873] &0.718 [0.692, 0.738] &0.391 [0.299, 0.468] & \\
\bottomrule
\end{tabular}
\caption{Confidence ranges of scores of translations on \texttt{newstest-18-head150} created by GA \textbf{without} knowledge of the reference in the fitness function, using other hypotheses and MBR decoding instead. See \cref{tab:ga_mbr}.}\label{tab:ga_mbr_sig}
\end{table*}

\begin{table*}[!htp]\centering
\scriptsize

\begin{tabular}{lrrrrrr}\toprule
&\textbf{ChrF} &\textbf{BLEU} &\textbf{CMT20} &\textbf{CMT21-mqm} &\textbf{CMTH22} \\\midrule
Reranking scores &0.575 [0.544, 0.607] &0.310 [0.268, 0.355] &0.598 [0.500, 0.681] &0.042 [0.040, 0.044] &0.560 [0.457, 0.653] \\
GA scores &0.575 [0.544, 0.605] &0.295 [0.253, 0.338] &0.626 [0.541, 0.699] &0.043 [0.041, 0.045] &0.541 [0.436, 0.630] \\
p-value for GA>reranking &0.505 &0.957 &0.004 &0 &0.941 \\
& & & & & \\ \midrule
&\textbf{QE} &\textbf{COMET22} &\textbf{BLEURT} &\textbf{UniTE} &\textbf{} \\ \midrule
Reranking scores &0.362 [0.302, 0.413] &0.852 [0.832, 0.869] &0.725 [0.700, 0.747] &0.368 [0.269, 0.453] & \\
GA scores &0.420 [0.365, 0.465] &0.859 [0.840, 0.874] &0.717 [0.693, 0.738] &0.394 [0.304, 0.471] & \\
p-value for GA>reranking &0 &0.008 &0.985 &0.006 & \\
\bottomrule
\end{tabular}
\caption{P-values for QE+CMT20+BLEU configuration being significantly better after GA compared to simple reranking  with the same objective function. We see that COMET22 and UniTE scores, which are held-out and we consider them more trustworthy, are significantly better when using GA.}
\label{tab:sig}
\end{table*}

\end{document}